\documentclass[runningheads]{llncs}
\usepackage{graphicx}

\usepackage{tikz}
\usepackage{comment}
\usepackage{amsmath,amssymb} 
\usepackage{color}
\usepackage{booktabs}
\usepackage{centernot}
\usepackage{mathtools}
\usepackage{stmaryrd}
\usepackage{siunitx}
\usepackage{multirow}
\usepackage{caption}
\usepackage{hyperref}
\newcommand{\el}[1]{#1~\textit{et al.}}
\newsavebox\CBox
\def\textBF#1{\sbox\CBox{#1}\resizebox{\wd\CBox}{\ht\CBox}{\textbf{#1}}}
\newcommand{\bestscore}[1]{\textcolor{black}{\textBF{#1}}}

\makeatletter
\newcommand{\xMapsto}[2][]{\ext@arrow 0599{\Mapstofill@}{#1}{#2}}
\def\Mapstofill@{\arrowfill@{\Mapstochar\Relbar}\Relbar\Rightarrow}
\makeatother

\makeatletter

\newcommand{\Rmnum}[1]{\expandafter\@slowromancap\romannumeral #1@}
\makeatother

\usepackage[accsupp]{axessibility}  

\begin{document}
\pagestyle{headings}
\mainmatter
\def\ECCVSubNumber{7210}  

\title{Animation from Blur: Multi-modal Blur Decomposition with Motion Guidance} 

\titlerunning{Multi-modal Animation from Blur}
\author{Zhihang Zhong\inst{1,3} \and
Xiao Sun\inst{2} \and
Zhirong Wu\inst{2} \and
Yinqiang Zheng\inst{1} \and
Stephen Lin\inst{2} \and
Imari Sato\inst{1,3}}
\authorrunning{Z. Zhong et al.}
%
\institute{The University of Tokyo, \email{zhong@is.s.u-tokyo.ac.jp} \and
Microsoft Research Asia
\and
National Institute of Informatics}
\maketitle

\begin{abstract}
We study the challenging problem of recovering detailed motion from a single motion-blurred image. Existing solutions to this problem estimate a single image sequence without considering the motion ambiguity for each region. Therefore, the results tend to converge to the mean of the multi-modal possibilities.
In this paper, we explicitly account for such motion ambiguity, allowing us to generate multiple plausible solutions all in sharp detail.
The key idea is to introduce a motion guidance representation, which is a compact quantization of 2D optical flow with only four discrete motion directions. 
Conditioned on the motion guidance, the blur decomposition is led to a specific, unambiguous solution by using a novel two-stage decomposition network.
We propose a unified framework for blur decomposition, which supports various interfaces for generating our motion guidance, including human input, motion information from adjacent video frames, and learning from a video dataset. Extensive experiments on synthesized datasets and real-world data show that the proposed framework is qualitatively and quantitatively superior to previous methods, and also offers the merit of producing physically plausible and diverse solutions.
Code is available at \href{https://github.com/zzh-tech/Animation-from-Blur}{https://github.com/zzh-tech/Animation-from-Blur}.
\end{abstract}

\section{Introduction}
\label{sec:introduction}
Motion blur appears in an image when the recorded scene undergoes change during the exposure period of the camera. Although such blur may impart a dynamic quality to a photograph, it is often desirable to invert this blur to produce a sharper image with clearer visual detail. Conventionally, this deblurring task is treated as a one-to-one mapping problem, taking a motion blurred image as input and producing a single output image corresponding to a single time instant during the exposure. Recently, attention has been drawn to a more challenging problem of extracting an image sequence~\cite{jin2018learning} instead, where the images correspond to multiple time instances that span the exposure period, thus forming a short video clip from the blurred motion.

\begin{figure}[!t]
\centering
    \includegraphics[width=.9\linewidth]{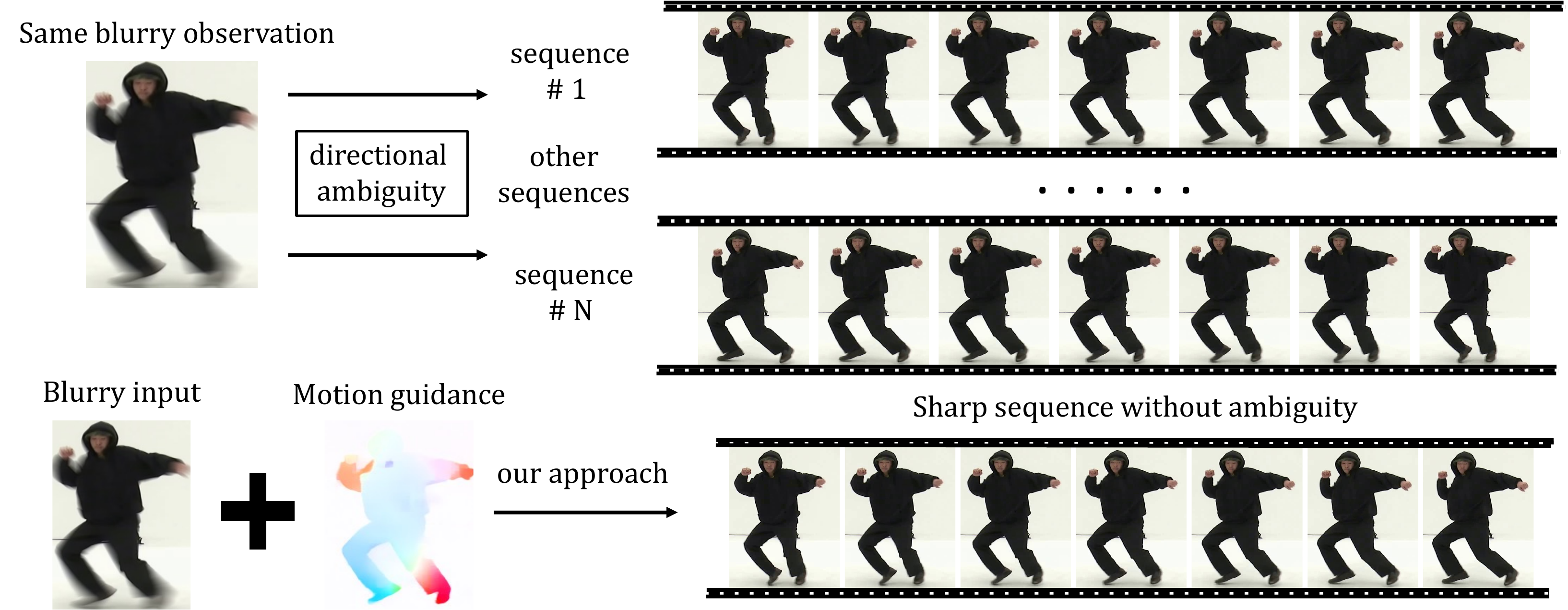}
    \caption{This paper studies the challenging problem of recovering the sharp image sequence from blurry input caused by relative motion. We first identify the fundamental directional motion ambiguity in this problem, which prevents the network to learn effectively. In order to solve this ambiguity, we propose to use motion as a guidance for conditional blur decomposition. Our approach demonstrates strong performance for recovering sharp visual details in motions.}
    \label{fig:teaser}
\end{figure}

Blur decomposition from a single blurry image faces the fundamental problem of motion ambiguity~\cite{purohit2019bringing}. Each independent and uniform motion blurred region in an image can correspond to either a forward or a backward motion sequence, both of which are plausible without additional knowledge. With multiple motion blurred regions in an image, the number of potential solutions increases exponentially, with many that are physically infeasible. For example, in Fig.~\ref{fig:teaser}, there exists multiple human dance movements that can correspond to the same observed blurry image, since the movement of each limb of the dancer can be independent. However, existing methods for blur decomposition are designed to predict a single solution among them. Moreover, this directional ambiguity brings instability to the training process, especially when the motion pattern is complex. As a result, this ambiguity, when left unaddressed as in current methods, would lead to poorly diversified and low-quality results.

In this work, we introduce a motion guidance representation to tackle the inherent directional motion ambiguity. The motion guidance is an optical flow representation quantized into four major quadrant directions, describing the motion field only roughly. Given the input blurry image, conditioned on the compact motion guidance, the blur decomposition now becomes a nearly deterministic one-to-one mapping problem without directional ambiguity. 
Empirically, we find that the decomposition network shows significantly better training convergence with this conditioning on an additional guidance input.

Given the blurry image and additional motion guidance as inputs, we propose a two-stage blur decomposition network to predict the image sequence. The first stage expands the blurry image into an image sequence based on the motion guidance, and the second stage refines the visual details in a residual fashion to generate high-quality images.
Our unified framework supports various interfaces for motion guidance acquisition:
1) Through a guidance predictor network learned from a dataset. The guidance predictor network is a VAE-GAN~\cite{larsen2016autoencoding,zhu2017multimodal} which learns a multi-modal distribution over plausible guidances. During inference, given an input blurry image together with sampled noises, the predictor can produce multiple guidance maps. 2) By additional information of dense optical flow computed from temporal frames. When the blurry image input is captured and sampled from a video sequence. The optical flow field between the blurry input and its adjacent frame can be used to compute the motion guidance. This motion guidance reflects the actual motion direction. 3) Through user input. Since the guidance is simple and compact, it can be annotated by outlining the region contours and their motion directions interactively. 

To train our model, we synthesize blurry images from high speed videos following the pipeline of REDS~\cite{nah2019ntire}. Specifically, we validate the performance of our model on human dance videos from Aist++~\cite{li2021ai} which only contain motion blur caused by human movement from a static camera, and general scene videos (GOPRO~\cite{nah2017deep}, DVD~\cite{su2017deep}), which are dominated by camera motion. Our approach provides a significant qualitative and quantitative improvement over previous methods by introducing a novel motion guidance representation to address the fundamental directional ambiguity in blur decomposition. Furthermore, due to the compactness of motion guidance representation, our unified framework may only need to be trained once, while supporting various decomposition scenarios under different modalities. The motion guidance obtained through different interfaces and their corresponding decomposition results reflect physically plausible and diverse solutions from our multi-modal framework. 

\section{Related Works}
\label{sec:related_works}
In this section, we review the related literature on image and video deblurring, blur decomposition, as well as multi-modal image translation. 

\subsection{Deblurring}
\label{sec:related_deblurring}
Deblurring refers to the task of estimating a sharp image from a blurry input, where the blur is often caused by scene or camera motion.
Traditional deblurring methods model the blur as a blur kernel operating on a sharp image via a convolution operation. A number of useful priors have been proposed to infer the latent sharp image and the corresponding blur kernel, such as total variation~\cite{chan1998total}, hyper-Laplacian~\cite{krishnan2009fast}, image sparsity~\cite{levin2009understanding}, and $l_0$-norm gradient~\cite{xu2013unnatural}. Recently, image and video deblurring has benefited from the advancement of deep convolution neural networks (CNNs). A coarse-to-fine pyramid CNN structure is widely used for the single-image deblurring task and achieves impressive performance~\cite{nah2017deep,tao2018scale,zhang2019deep}. Also, generative adversarial networks (GANs) have been adopted to improve the perceptual quality of deblurring results~\cite{kupyn2018deblurgan,kupyn2019deblurgan}.
Temporal dependency across adjacent frames is another source of information which could be utilized to recover the sharp image~\cite{su2017deep,wang2019edvr}. Recurrent architectures are shown be effective at exploiting such temporal information~\cite{hyun2017online,zhou2019spatio,nah2019recurrent,zhong2020efficient}. Extensive studies in these works demonstrate that deep neural networks are able to approximate blur kernels well in an implicit way. 

\subsection{Blur Decomposition}
\label{sec:related_blur_decompsition}
Blur decomposition attempts to recover the full image sequence from a blurry input caused by object and camera motion during the exposure time. \cite{jin2018learning} are the first to tackle this problem and propose a pairwise order-invariant loss to improve convergence in training. \cite{purohit2019bringing} present a method that extracts a motion representation from the blurry image through self-supervised learning, and then feeds it into a recurrent video decoder to generate a sharp image sequence. \cite{argaw2021restoration} utilize an encoder-decoder structure with a spatial transformer network to estimate the middle frame and other frames simultaneously with a transformation consistency loss. Assuming that the background is known, DeFMO~\cite{rozumnyi2021defmo} embeds a blurred object into a latent space, from which it is rendered into an image sequence that is constrained to be sharp, time-consistent, and independent of the background. 

When the input is a blurry video, \cite{jin2019learning} design a cascade scheme, \textit{i.e.}, deblurring followed by interpolation, to tackle the problem. To avoid errors introduced in the first stage, \el{Shen} propose BIN~\cite{shen2020blurry} with a pyramid module and an inter-pyramid recurrent module to jointly estimate the latent sharp sequence. \cite{argaw2021motion} achieve blurry video decomposition by first estimating the optical flow and then predicting the latent sequence by warping the decoded features. In addition, methods for blur decomposition also consider exploiting high-frequency temporal information from event cameras~\cite{pan2019bringing,lin2020learning}.

None of the existing methods address the fundamental ambiguity that exists with motion directions. 
We are the first to address this by conditioning the decomposition process via a novel motion guidance approach. We also design a flexible blur decomposition network, which can produce diverse decomposition results using guidance from the proposed interfaces for different modalities.

\subsection{Image-to-Image/Video Translation}
\label{sec:related_translation}

Our work is related to image-to-image translation networks~\cite{isola2017image} with applications for image synthesis~\cite{park2019semantic} and style transfer~\cite{zhu2017unpaired}. GANs~\cite{goodfellow2014generative,karras2019style} and VAEs~\cite{kingma2013auto} are two popular approaches for training generative models. Both models can easily be conditioned by feeding additional inputs, such as in conditional VAE~\cite{sohn2015learning} and infoGAN~\cite{chen2016infogan}. Blur decomposition can be formulated as a similar image-to-image translation problem. However, the inherent motion ambiguity in the blurry image prevents the model from converging to a single optimal solution. We thus introduce a new motion guidance representation to disambiguate the motion directions.

Multi-modal image translation is a promising direction to generate a distribution of results given a single input. Approaches along this direction combine VAE and GAN, so that stochastic variables in VAE could sample from a distribution, while GAN is used to encourage realistic generations. Notable works include 
VAE-GAN~\cite{larsen2016autoencoding} and BicycleGAN~\cite{zhu2017multimodal}. Our guidance predictor follows VAE-GAN~\cite{larsen2016autoencoding} to generate multi-modal motion guidance. We note that this technical element is not among the contributions of our paper.

Learning to animate a static image is a fascinating application. Learning-based methods either predict the motion field~\cite{holynski2021animating,endo2019animating} which is later used to generate videos, or directly predict the next frames~\cite{zhang2020dtvnet,xue2016visual}. Our work focuses on animating a specific kind of input image with motion blur, which is physically informative to recover the image sequence. 

\section{Methodology}
\label{sec:methodology}

\textbf{The Blur Decomposition Problem.}
A blurred image can be considered as an average of successive relatively sharp images over the exposure time 
\begin{equation}
    I_b = \frac{1}{T}\int_{0}^{T}I^tdt,
\end{equation}
which can be approximately expressed in the following discrete form when $T$ is large enough, \textit{i.e.}, $I^{1} \cdots I^{T}$ is a high-frame-rate sequence
\begin{equation}
    \dfrac{1}{T}(I^1 + ... + I^T) = I_b.
    \label{eq.blur_syn}
\end{equation}
Note that this image averaging process is simulated in an approximately linear space through applying inverse gamma calibration to RGB images~\cite{nah2017deep,nah2019ntire}. Our goal is to invert this blurring process to estimate a finite sequence of sharp images which are uniformly distributed over the exposure time
\begin{equation}
    I_b \xmapsto{\mathbf{D}} \mathbf{I} = \{I^t, t\in 1, \cdots T \}.
    \label{eq.decomposition}
\end{equation}
This is a highly ill-posed problem because given the blurry image $I_b$, there are infinitely many solutions per pixel (u, v) among the sharp images if these sharp images are treated as independent
\begin{equation}
    \dfrac{1}{T}(I^1(u,v) + ... + I^T(u,v))  = I_b(u,v).
    \label{eq.ambiguity_pixel}
\end{equation}
However, assuming the images exist in succession over a short period of time and that the pixels are rarely affected by saturation and occlusions/disocclusions, then the following holds true: the pixels in these images are highly correlated and their dependencies can be described by the optical flow $F^t$ of a sharp image $I^t$ to its next frame $I^{t+1}$,
\begin{equation}
    I^t(u,v) = I^{t+1}(F^t(u,v)).
    \label{eq.flow_pixel}
\end{equation}
Supposing that we are given the optical flow fields for all frames in the target sequence $\mathbf{F} = \{F^t, t\in 1, \cdots T-1 \}$,
Equations~\ref{eq.ambiguity_pixel} and~\ref{eq.flow_pixel} form a linear system which can be solved with a unique solution. In other words, the ground truth motions between sharp images resolve the ambiguity problem in Equation~\ref{eq.decomposition}.

\begin{figure}[!t]
  \centering
  \includegraphics[width=.85\linewidth]{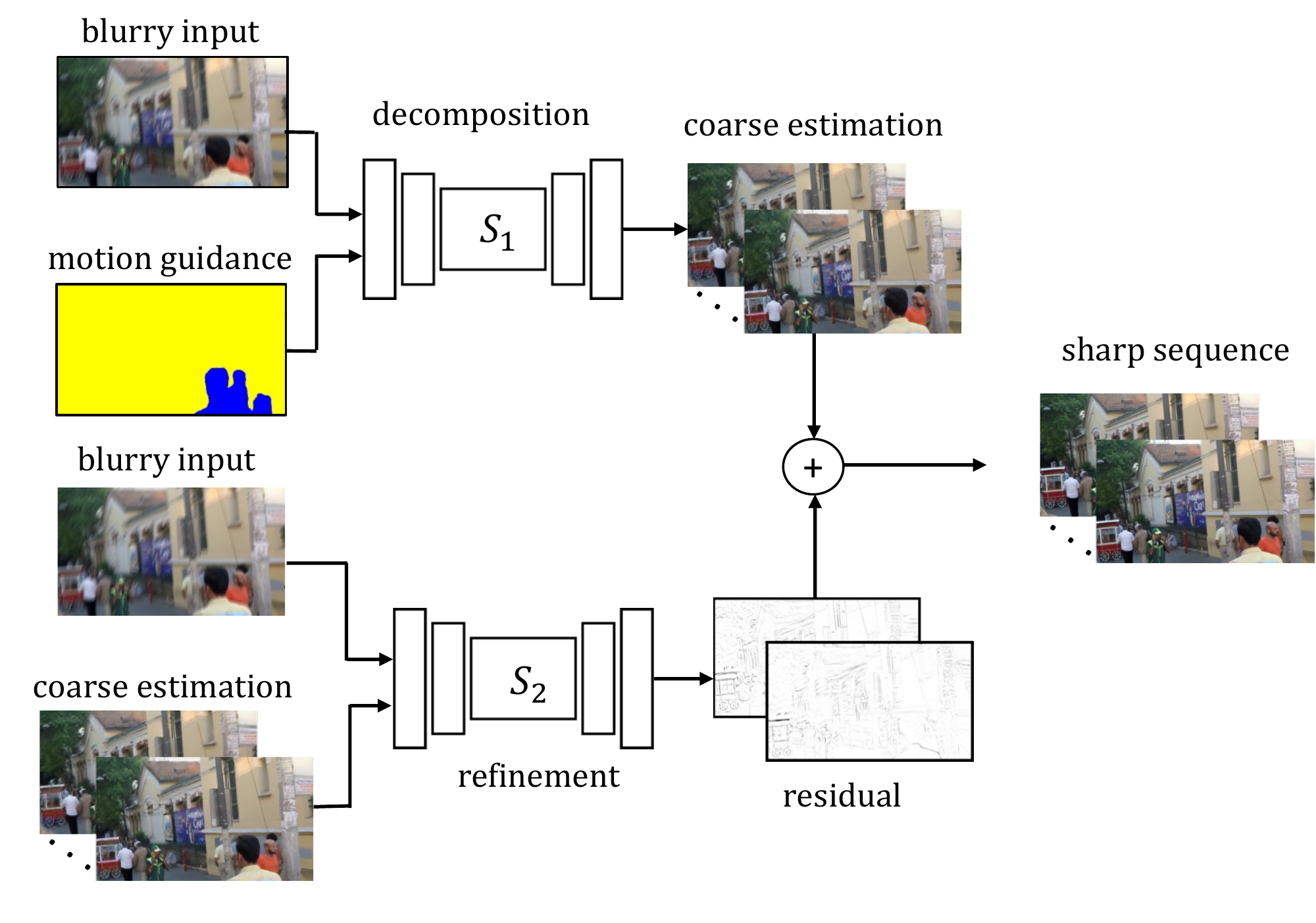}
  \caption{\textbf{Architecture of our motion guided blur decomposition network.} It consists of two successive stages. The first stage $S_1$ concatenates the blurry image $I_b$ and motion guidance $G$ as input and outputs a rough image sequence. The second stage $S_2$ then refines the visual details in a residual fashion.}
  \label{fig:decomposer}
\end{figure}

\noindent \textbf{Motion-guided Decomposition.}
We propose to solve the decomposition problem in Equation~\ref{eq.decomposition} in two steps. The first step estimates the motions $\mathbf{F}$ during the exposure time, \textit{e.g.}, by learning a motion predictor $\mathcal{P}$ for the blurry image, \begin{equation}
    I_b \xmapsto{\mathcal{P}} \mathbf{\hat{F}}.
    \label{eq.predictor}
\end{equation}
The second step learns a motion guided decomposition network $\mathcal{S}$ that takes both the blurry image and an estimated motion guidance as input to predict the sharp image sequence,
\begin{equation}
    (I_b, \mathbf{\hat{F}}) \xmapsto{\mathcal{S}} \mathbf{I} = \{\hat{I}^t, t\in 1, \cdots T \}.
    \label{eq.decomposition_flow}
\end{equation}
As a result, the ambiguity is explicitly decoupled from the decomposition network $\mathcal{S}$, and only exists in the motion predictor $\mathcal{P}$. 

Predicting the motion guidance with frame-wise optical flow would still be a difficult task. In practice, motion ambiguity will be complex, because the ambiguities of the independent moving regions in the image can be arbitrarily combined. However, we notice that the ambiguity mainly lies in the forward and backward directions of the motions, and hence the motion guidance does not need to be precise to frame-wise and continuous values to resolve the ambiguity. We therefore propose a compact motion guidance representation to replace the accurate optical flow, which enables the decomposition network to adapt to different input modalities. In the following, we describe the compact motion guidance representation in Sec.~\ref{sec:guidance}, the blur decomposition network conditioned on the motion guidance in Sec.~\ref{sec:decomposer}, and three distinct interfaces for acquiring the motion guidance in Sec.~\ref{sec:unified_interface}.

\subsection{Motion Guidance Representation}
\label{sec:guidance}
Given accurate dense optical flow sequence $\mathbf{F} = \{F^t, t\in 1, \cdots T-1 \}$, the blur decomposition problem can be solved without ambiguity. However, dense optical flow sequence is difficult to obtain and to predict accurately, and hence it may not be an ideal representation as a guidance.

We notice that the ambiguity for blur decomposition is in the motion directions. For example, if $\mathbf{F} = \{F^1, ... , F^{T-1}\}$ is one possible motion of the blurry image $I_b$ and the corresponding sharp image sequence is $\mathbf{I} = \{I^1, ..., I^T\}$, then there also exists another reverse motion $\mathbf{F_{\text{bac}}} = \{F_{\text{bac}}^{T-1}, ... , F_{\text{bac}}^{1}\}$ with the corresponding sharp image sequence $\mathbf{I_{\text{inv}}} = \{I^T,...,I^1 \}$ which leads to the same blurry image $I_b$, where $F_{\text{bac}}^{t}$ is the backward optical flow between $I^{t}$ and $I^{t+1}$. Providing a crude motion direction may be sufficient to resolve the ambiguity.

Motivated by this observation, we design a compact guidance representation by motion quantization. We first assume that the motion directions within the exposure time do not change abruptly. This is generally true when the shutter speed is not extremely slow compared with object motion. We thus use the aggregated flow to represent the motion pattern for the full sequence $\bar{F} = \sum_1^T F^t$. We further quantize the aggregated flow into four quadrant directions and an additional motionless class which takes flows below a certain magnitude. Empirically, we find four quadrant directions to be adequate for disambiguating motion directions. We denote the motion guidance as $G$.

\subsection{Motion Guided Blur Decomposition Network}
\label{sec:decomposer}

Given the blurry image and the motion guidance, the sharp image sequence is predicted via a blur decomposition network $\mathcal{S}$. Once the model is trained, it can be used to decompose a blurry image into different sharp image sequences simply by providing it with the corresponding motion guidance.
In this section, we illustrate the model architecture and training loss in detail. 

Training the network requires a dataset of triplet samples $(I_b, G, \mathbf{I})$, consisting of the blurry image $I_b$, the ground truth sharp image sequence $\mathbf{I}$, and motion guidance $G$ derived from $\mathbf{I}$. We follow the common practice~\cite{nah2017deep,nah2019ntire} of synthesizing a blurry image by accumulating sharp images over time according to Equation~\ref{eq.blur_syn}, which is implemented in linear space through inverse gamma calibration. We use a off-the-shelf optical flow estimator~\cite{teed2020raft} for deriving $\mathbf{F}$ and the guidance $G$.

Fig.~\ref{fig:decomposer} illustrates our two-stage workflow of $\mathcal{S}$. 
The first stage estimates a rough dynamic image sequence, and the second stage refines the visual details in a residual fashion.
Both networks adopt a similar encoder-decoder architecture. The architecture details can be found in the supplementary materials. The two-stage blur decomposition network outputs the sharp image sequence prediction $\mathbf{\hat{I}}$. We adopt commonly used $\mathcal{L}_2$ loss function for supervising the decomposition,
\begin{equation}
    \mathcal{L}_2 = \|\mathbf{I} - \mathbf{\hat{I}}\|_2^2.
\end{equation}
 
\subsection{Motion Guidance Acquisition}
\label{sec:unified_interface}
\begin{figure}[!t]
    \centering
   \includegraphics[width=\linewidth]{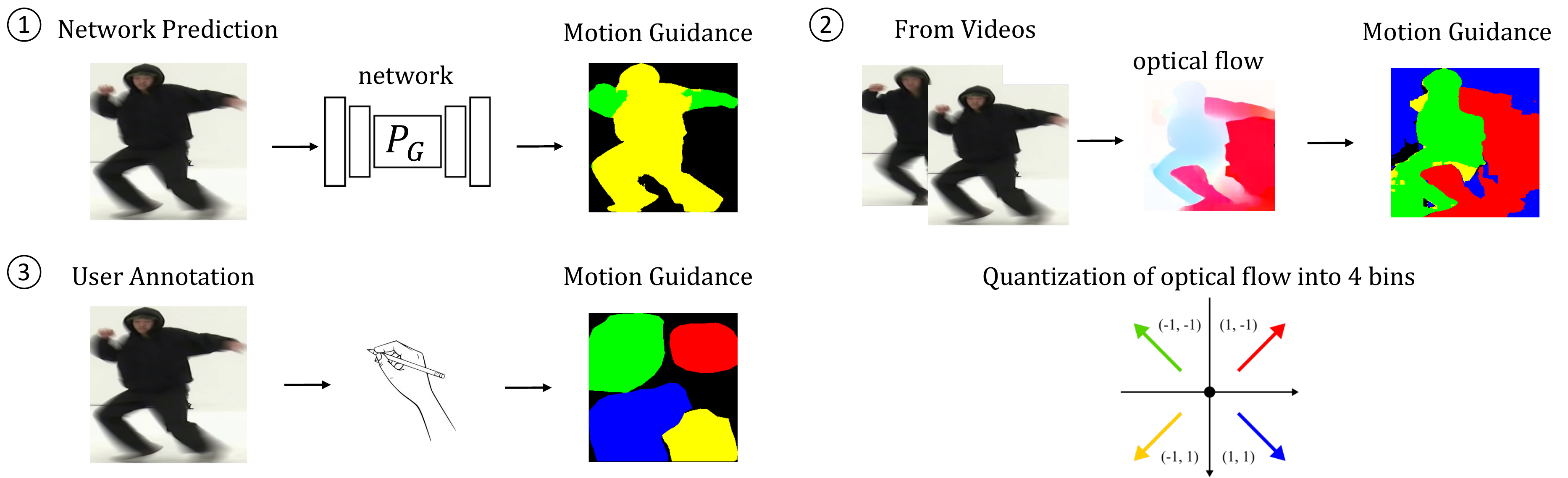}
   \caption{\textbf{Motion guidance acquisition.} We represent motion guidance as a quantized motion vector into 4 quadrant directions. Due to its compactness, the guidance could be obtained from a network, videos or user annotations.}
   \label{fig:mbd}
\end{figure}

We provide three interfaces to acquire the motion guidance: learning to predict the motion guidance, motion from video, and user input, as illustrated in Fig.~\ref{fig:mbd}.

\begin{figure*}[!t]
  \centering
  \includegraphics[width=\linewidth]{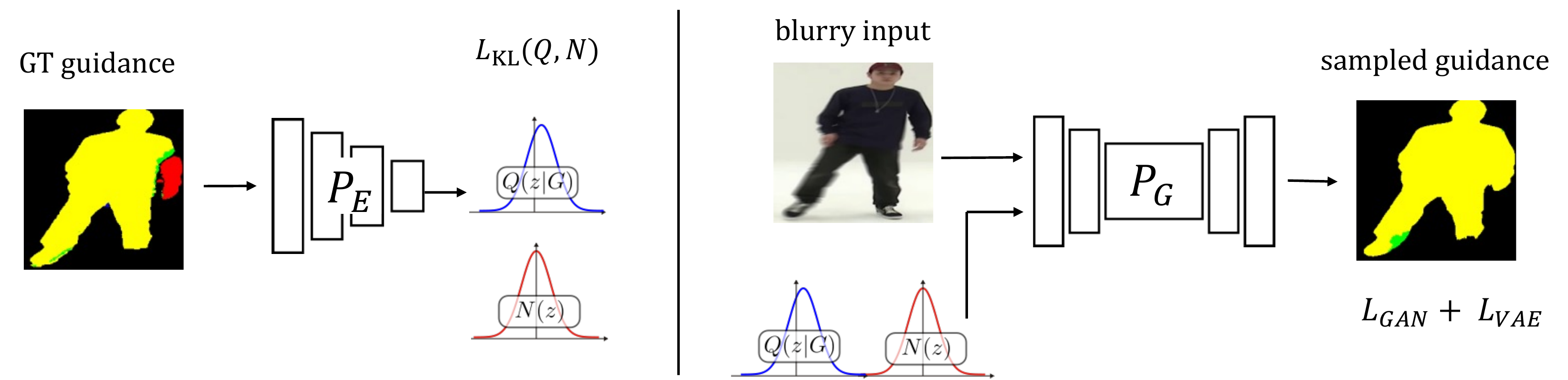}
  \caption{\textbf{Framework of the multi-modal motion guidance prediction network.} We mainly follow the cVAE-GAN~\cite{sohn2015learning,larsen2016autoencoding}, and adapt it to our problem to generate multiple physically plausible guidances from a blurry image.}
  \label{fig:predictor}
\end{figure*}

\noindent \textbf{Multi-modal Motion Prediction Network.}
The directional ambiguity now exists in the motion guidance representation. To account for the ambiguity, we train a multi-modal network to generate multiple physically plausible guidances given a blurry image. We follow the framework of cVAE-GAN~\cite{sohn2015learning,larsen2016autoencoding} for multi-modal image translation as shown in Fig.~\ref{fig:predictor}.

The guidance prediction network comprises an encoder $\mathcal{P}_E$ and a generator $\mathcal{P}_G$. The encoder $\mathcal{P}_E$ converts the ground truth guidance into a latent stochastic variable by $\mathbf{z} =\mathcal{P}_E(G)$, and the generator predicts the guidance given the latent code and the blurry input $\hat{G} = \mathcal{P}_G(\mathbf{z}, I_b)$. The latent variable $\mathbf{z}$ is considered to follow a Gaussian distribution $\mathcal{N}(0, 1)$, and this stochastic variable is used to model the directional distribution that exists in the motion space. At the testing phase, multi-modal motion predictions can be generated by randomly sampling the latent vector $\mathbf{z}$ from a Gaussian distribution.

The overall network is trained with a combination of a GAN loss $\mathcal{L}_{\text{GAN}}$ on the guidance predictions, a VAE loss $\mathcal{L}_{\text{VAE}}$, and a Kullback-Leibler (KL) divergence loss $\mathcal{L}_{\text{KL}}$ for the stochastic variable $\mathbf{z}$ as in~\cite{zhu2017multimodal},
\begin{equation}
\begin{split}
\mathcal{L}_{\text{guidance}} = &  \lambda_{1}\mathcal{L}_{\text{GAN}}(\mathcal{P}_E, \mathcal{P}_G) \\
+ & \lambda_{2} \mathcal{L}_{\text{VAE}}(\mathcal{P}_E, \mathcal{P}_G) + \lambda_{3} \mathcal{L}_{\text{KL}}( \mathbf{z} || \mathcal{N}(0,1)),
\end{split}
\end{equation}
where $\lambda_{1},\lambda_{2},\lambda_{3}$ are mixing coefficients for the losses. The architecture details can be found in the supplementary materials.

\noindent\textbf{Motion from Video.} We assume that the motion direction does not change abruptly in a short time. Thus, the motion direction in the blurry image can be approximated by the motion to the adjacent frames, if a video is provided. Hence, our method can be directly applied to video deblurring without any modification. In the experiment, our method is also compared with the state-of-the-art video deblurring methods and shows better performance.

\noindent\textbf{Human Annotation.} The compact quantized motion guidance provides a friendly interface to our blur decomposition network. Given a blurry image, the user can generate multiple plausible sharp video clips simply by drawing the outlines of the blurry regions and arbitrarily specifying their motion directions.

\section{Experiments}
\label{sec:experiments}
We show the details of the used datasets in Sec.~\ref{sec:dataset}, single image and video decomposition results in Sec.~\ref{sec:image_evaluation} and  Sec.~\ref{sec:video_evaluation}, real-world evaluation in Sec.~\ref{sec:realworld}, guidance robustness analysis in Sec.~\ref{sec:robustness}, as well as ablation study in Sec.~\ref{sec:ablation}.

\subsection{Datasets}
\label{sec:dataset}
Existing dataset for blur decomposition~\cite{shen2020blurry} introduces $1/3$ temporal overlaps between adjacent blurred frames. This violates the actual blur occurred in real world. We thus create datasets by our own where blurry frames have almost no temporal overlaps. One of our datasets for general scenes (GenBlur) consists of high-fps videos used by related works~\cite{jin2018learning,purohit2019bringing}, but with a better pipeline~\cite{nah2019ntire} to simulate the formation of blur.

\begin{figure*}[!t]
  \centering
  \includegraphics[width=\linewidth]{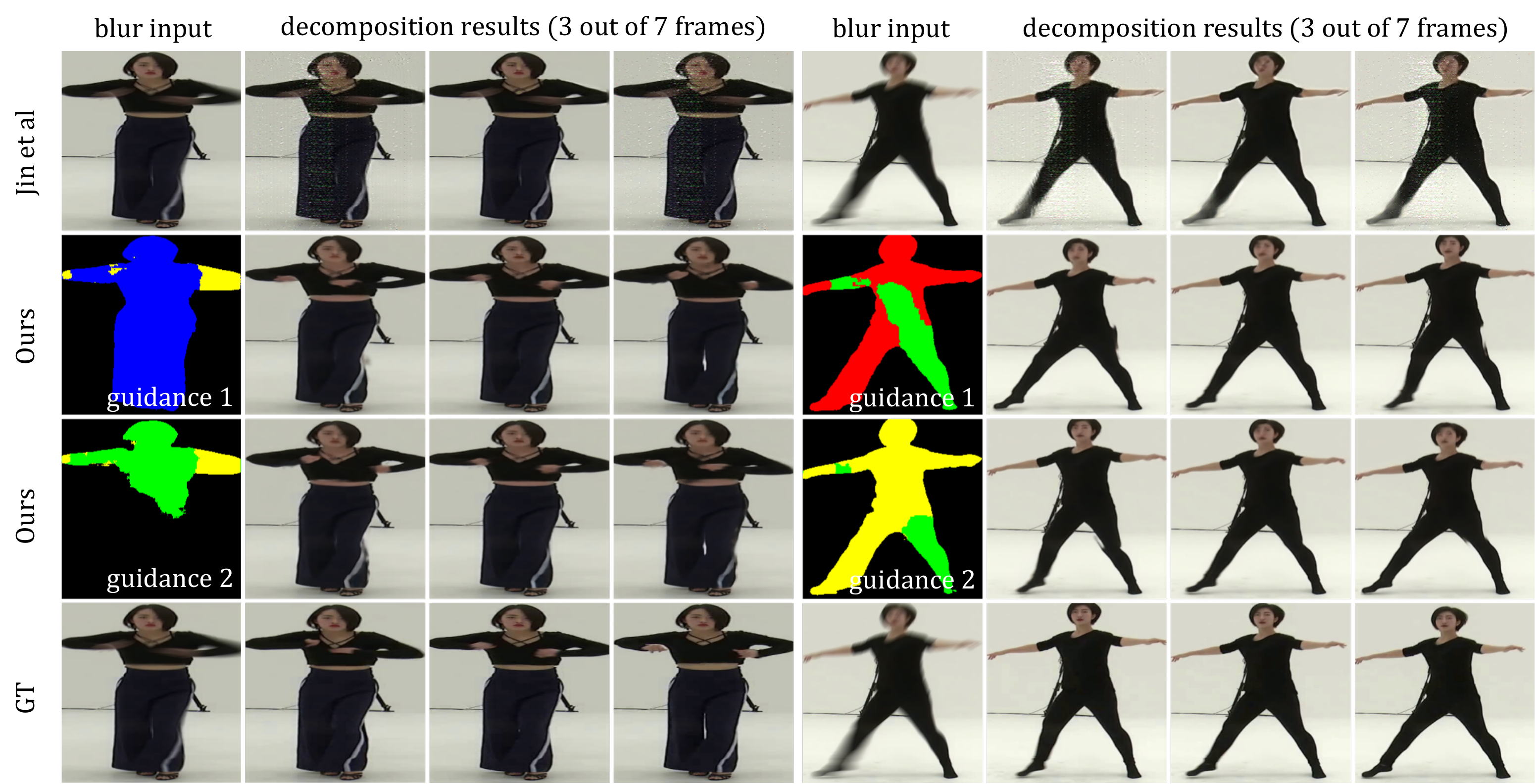}
  \caption{\textbf{Qualitative comparison with single image decomposition baseline method~\cite{jin2018learning}.} Given a blurry image, our method can generate multiple physically plausible motion guidance and recover distinct sharp image sequences based on each motion guidance. The baseline \el{Jin} fails, resulting in a sequence with little motion. Please pay attention to the motion on the hands of the left dancer and the legs of the right dancer.}
  \label{fig:comparison_image}
\end{figure*}

\noindent\textbf{GenBlur.}
We synthesize the GenBlur dataset using high-frame-rate (\SI{240}{fps}) videos from GOPRO~\cite{nah2017deep}, DVD~\cite{su2017deep}, and videos collected by ourselves. To suppress  noise and video compression artifact~\cite{nah2019recurrent}, the image resolution is uniformly down-sampled to 960$\times$540. We follow a widely used blur synthesis technique proposed in ~\cite{nah2019ntire}, that first employs an off-the-shelf
CNN~\cite{huang2020rife} to interpolate frames into a much higher fps (\SI{240}{fps} $\to$ \SI{7680}{fps}) to avoid unnatural spikes or steps in the blurred trajectory~\cite{wieschollek2017learning}. In the interpolated videos, sets of 128 frames (non-overlapping) are averaged in linear space through inverse gamma calibration to synthesize blurred images. Instead of using all the 128 frames as ground truth clear images, we evenly sample 7 images among them to keep consistent with the previous work~\cite{jin2018learning}. There are 161 and 31 videos for the train and test set, respectively. This dataset consists mainly of urban scenes, and the motion is dominated by camera ego-motion.

\noindent \textbf{B-Aist++.}
The problem arising from directional ambiguity is particularly severe when there are multiple independent motions in the image. We thus synthesize another dataset specifically to highlight this issue by using videos from a dance dataset Aist++~\cite{li2021ai} which contains complex human body movements by professional dancers. We use the same pipeline as for GenBlur to synthesize blurry images and corresponding sharp image sequences. The synthesized blurry dataset is denoted as B-Aist++. There are 73 and 32 videos for the train and test set. The images are in resolution of 960$\times$720. This dataset contains complex human motion and the camera is stationary.

\subsection{Blurry Image Decomposition}
\label{sec:image_evaluation}

\noindent\textbf{Qualitative Evaluation.}
We qualitatively compare our result with the state-of-the-art single image based blurry decomposition method~\cite{jin2018learning} in Fig.~\ref{fig:comparison_image}. B-Aist++ is used for comparison because of its complex directional ambiguities. In the second and third rows, we demonstrate diversity in our blur decomposition results through the use of different motion guidance sampled from our motion predictor $\mathcal{P}$. Ground-truth is presented in the last row.

Existing methods such as~\cite{jin2018learning} are unable to resolve the directional ambiguity in motion blur. For example, for the blur in the second dancer's legs, they cannot determine whether the legs are being spread out or drawn in. Consequently, in the case of data with high directional ambiguity such as B-Aist++, the moving range of the generated sharp frames is limited. In contrast, our multi-modal method incorporates directional guidance to remove ambiguity, leading to multiple coherent natural motions that look physically plausible. To better perceive the temporal variation in the decomposition results, we strongly recommend the reader to check out the videos in our supplementary materials.

\begin{table}[!t]
  \setlength{\tabcolsep}{6pt}
  \centering
  \footnotesize
  \caption{\textbf{Quantitative evaluation of single blurry image decomposition.} For our method, we predict multiple motion guidance from our guidance predictor network.  $P_{\#}$ denotes we evaluate $\#$ number of plausible decomposition results for each input, and choose the best case. The results of Jin~\textit{et al.}~\cite{jin2018learning} represent the best performance calculated as using either the forward or reverse outputs, following the original paper. Our approach outperforms Jin~\textit{et al.}~\cite{jin2018learning} by a large margin even with a single sampling.
  }
  \label{tab:comparison_image}
  \begin{tabular}{lcccccc}
    \toprule
    Dataset & \multicolumn{3}{c}{B-Aist++} & \multicolumn{3}{c}{GenBlur} \\
    Method & PSNR $\uparrow$ & SSIM $\uparrow$ & LPIPS $\downarrow$ & PSNR $\uparrow$ & SSIM $\uparrow$ & LPIPS $\downarrow$ \\
    \midrule
    Jin~\textit{et al.}~\cite{jin2018learning} & 17.01 & 0.540 & 0.192 & 20.88 & 0.621 & 0.283\\
    Ours ($\mathcal{P}_1$) & 19.97 & 0.860 & 0.089 & 23.41 & 0.737 & 0.267\\
    Ours ($\mathcal{P}_3$) & 22.44 & 0.898 & 0.068 & 23.56 & 0.740 & 0.263\\
    Ours ($\mathcal{P}_5$) & \bestscore{23.49} & \bestscore{0.911} & \bestscore{0.060} & \bestscore{23.61} & \bestscore{0.741} & \bestscore{0.260}\\
    \bottomrule
  \end{tabular}
\end{table}

\noindent\textbf{Quantitative Evaluation.}
We report quantitative comparison results on B-Aist++ and GenBlur in Table~\ref{tab:comparison_image}. Following common practice, PSNR, SSIM and LPIPS~\cite{zhang2018unreasonable} are used as evaluation metrics. Our method can generate multiple decomposition results by sampling multiple motion guidances from the motion predictor $\mathcal{P}$. In the table, $\mathcal{P}_{\#}$ denotes that $\#$ motion guidances are sampled from the predictor and the best result among the samples is reported. It can be seen that our method outperforms Jin~\textit{et al.}~\cite{jin2018learning} by a large margin and more samples leads to better best results.

\subsection{Blurry Video Decomposition}
\label{sec:video_evaluation}

\begin{figure*}[!t]
  \centering
  \includegraphics[width=\linewidth]{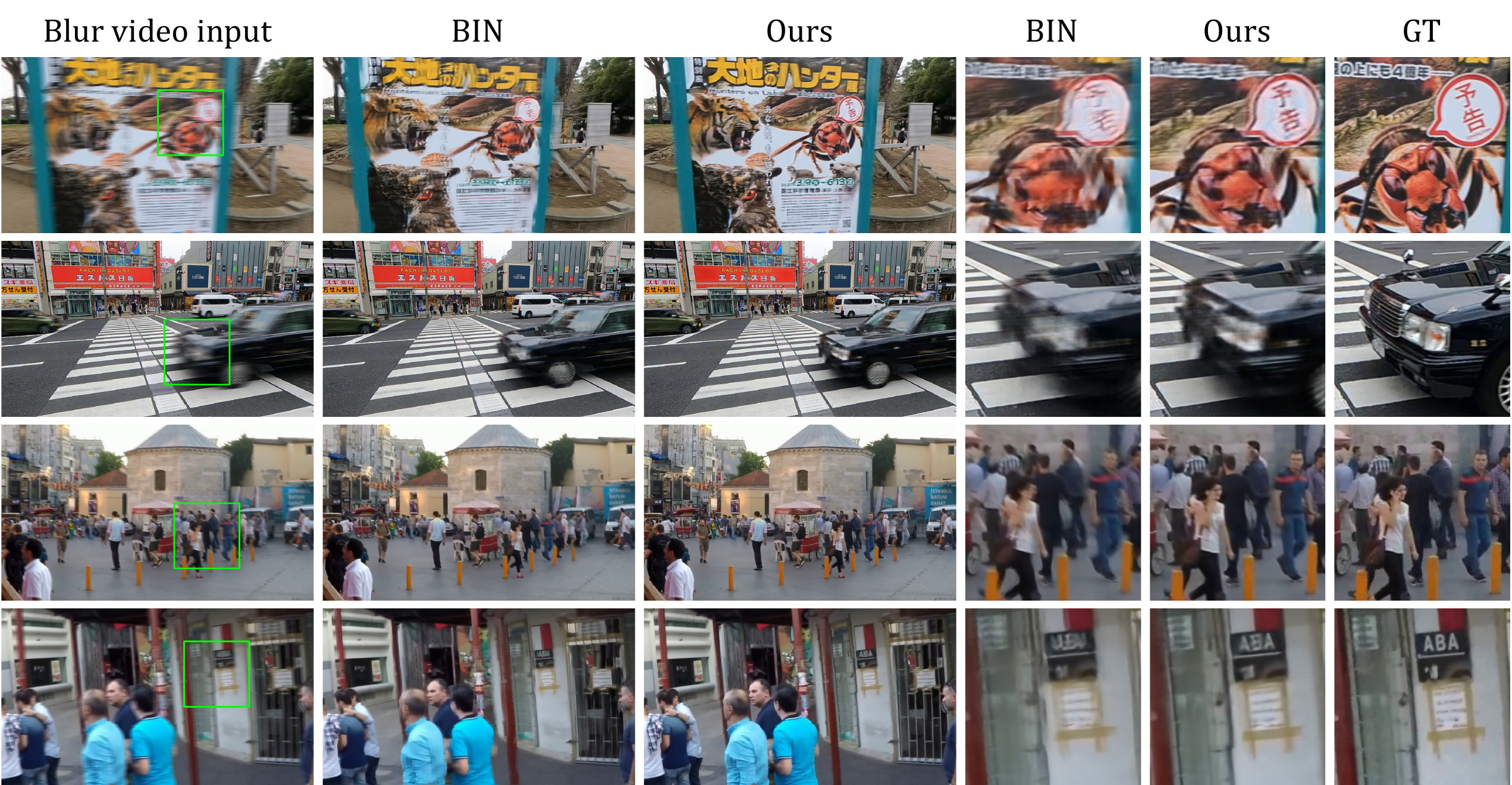}
  \caption{\textbf{Qualitative comparison with blurry video decomposition baseline method BIN~\cite{shen2020blurry}.} Our approach uses the guidance calculated from the input video itself. For close up investigation, our method recovers much sharper visual details than the baseline method, \textit{e.g.}, the text in the blurry input.}
  \label{fig:comparison_video}
\end{figure*}

\begin{table}[!t]
\setlength{\tabcolsep}{3pt}
  \centering
  \footnotesize
  \caption{\textbf{Quantitative evaluation of blurry video decomposition.} The motion guidance for our method is obtained from the optical flow in the input blurry video. Our method outperforms the baseline BIN~\cite{shen2020blurry} by a large margin of $1.16$ dB and $2.22$ dB absolute PSNR on B-AIST++ and GenBlur respectively.
  }
  \label{tab:comparison_video}
  \begin{tabular}{lcccccc}
    \toprule
    Dataset & \multicolumn{3}{c}{B-Aist++} & \multicolumn{3}{c}{GenBlur} \\
    Method & PSNR $\uparrow$ & SSIM $\uparrow$ & LPIPS $\downarrow$ & PSNR $\uparrow$ & SSIM $\uparrow$ & LPIPS $\downarrow$ \\
    \midrule
    BIN~\cite{shen2020blurry} & 22.84 & 0.903 & 0.068 & 24.82 & 0.805 & 0.157\\
    Ours (guidance from video) & \bestscore{24.03} & \bestscore{0.911} & \bestscore{0.067} & \bestscore{27.04} & \bestscore{0.858} & \bestscore{0.122} \\
    \bottomrule
  \end{tabular}
\end{table}

\noindent\textbf{Qualitative Evaluation.}
As explained in Sec.~\ref{sec:unified_interface}, our method can be directly applied to video based decomposition without modification. The motion guidance used for our decomposition network is quantized from the optical flow between adjacent frames.

Fig.~\ref{fig:comparison_video} presents a visual comparison between our result and the state-of-the-art video-based method BIN~\cite{shen2020blurry} on general scenes (GenBlur). Our decomposition results surpass those of the video-based model, with much clearer details. It is worth noting that since the adjacent frames themselves are blurred, the estimated optical flows are not accurate. However, due to the effective direction augmentation and the need for only coarse directional guidance, the learned decomposition network is robust and effective. Also, please see the videos in our supplementary for better perception.

\noindent\textbf{Quantitative Evaluation.}
When comparing with the video-based method BIN~\cite{shen2020blurry}, we use motion guidance estimated from adjacent frames using the off-the-shelf flow estimator~\cite{teed2020raft}, denoted as \textit{vid.} in Table.~\ref{tab:comparison_video}. Our method is clearly superior to~\cite{shen2020blurry} in terms of all metrics.

\subsection{Real-world Evaluation}
\label{sec:realworld}
We further captured real-world blurry image for validating the generalization ability of the proposed method. Because the blurry image is captured in real world, no ground truth sharp image sequences can be used for quantitative evaluation. We display the visual results on real-world data when a user provides directional guidance in Fig.~\ref{fig:real}. The results demonstrate that our method can generalize well on real data. We also show real-world evaluations with the guidance prediction network in the appendix.

\begin{figure}[!t]
  \centering
  \includegraphics[width=.9\linewidth]{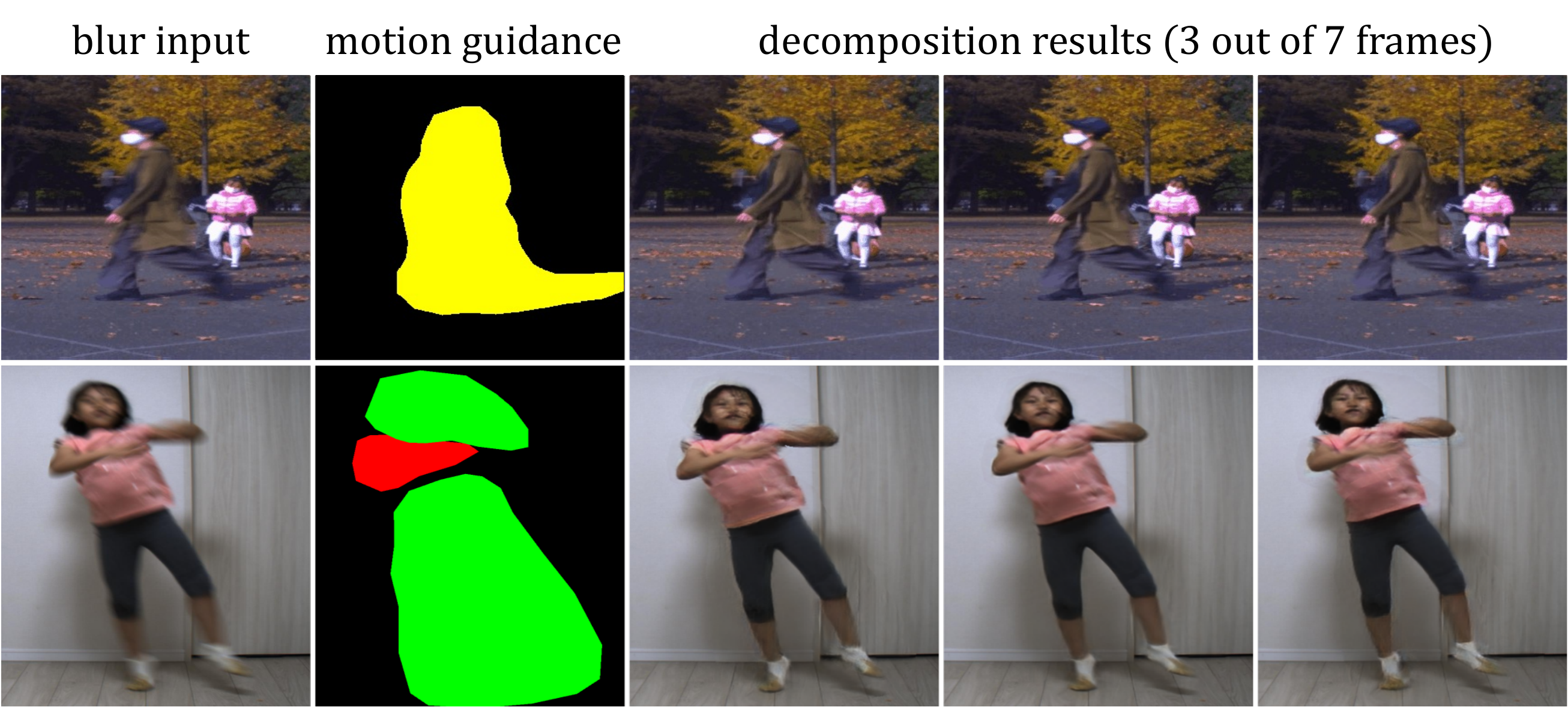}
  \caption{\textbf{Results on real-world captured blurry image.} We manually provide the motion guidance and recover the image sequence using our decomposition network. Although trained on synthetic data, our model successfully generalizes on real-world data. Please notice the relative position between the door frame and the legs of the girl in the bottom. }
  \label{fig:real}
\end{figure}

\subsection{Robustness of motion guidance}
\label{sec:robustness}
\begin{figure}[ht]
    \centering
    \includegraphics[width=\linewidth]{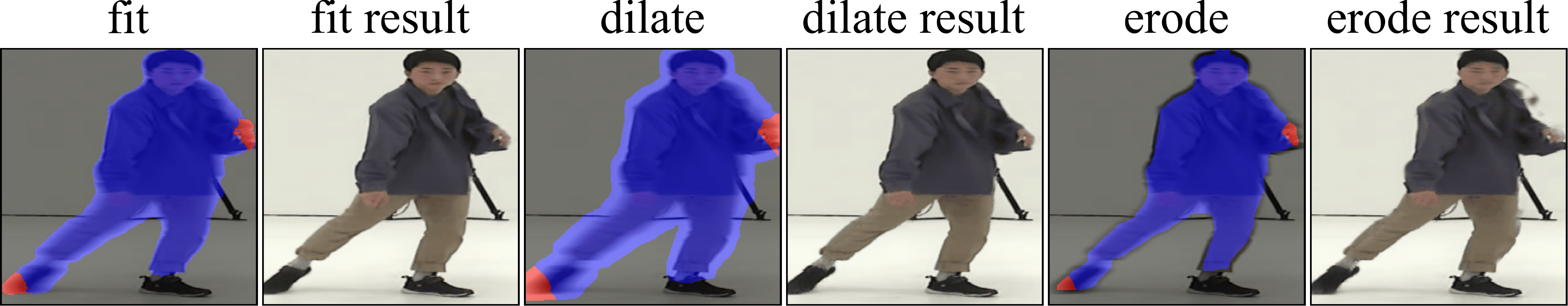}
    \caption{\textbf{The influence of incorrect prediction or human annotation.} Our approach may tolerate guidance errors in the sharp regions.}
    \label{fig:robustness}
\end{figure}

The errors of motion guidance may come from two sources: one by incorrect prediction or human annotation, one by optical flow quantization. For the first case, the model may tolerate guidance errors in the sharp regions. We demonstrate this by showing results using fit, dilated, and eroded guidance in Fig.~\ref{fig:robustness}. The decomposition model may identify the pixels are non-blurry and preserve the original details, no matter what guidance prediction is given (dilate result). The model cannot recover a blurry region and may introduce artifacts if it is not provided with a meaningful guidance (erode result). For the second case, please refer to the ablation experiment in the supplementary material, which shows that quantization into 4 bins is sufficient to achieve good performance.

\subsection{Ablation Studies}
\label{sec:ablation}
\begin{figure}[t]
\centering
\begin{minipage}[h]{0.48\textwidth}
\centering
  \includegraphics[width=\linewidth]{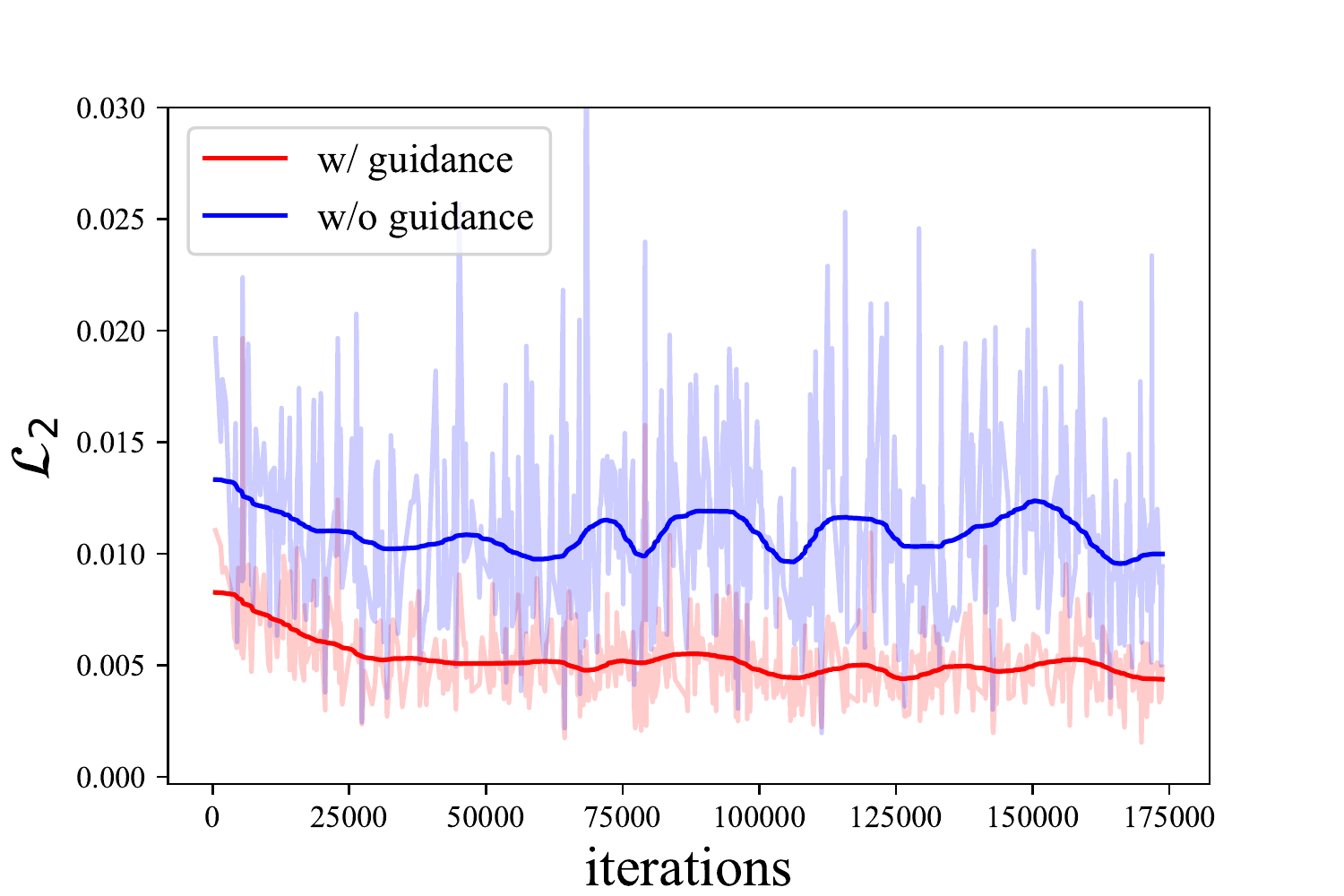}
  \caption{\textbf{Training curves for the decomposition network with and without motion guidance.} With extra guidance input to disambiguate the motion direction, the network is able to fit to the data a lot easier.}
  \label{fig:learning_curve}
\end{minipage}
\hfill
\begin{minipage}[h]{0.48\textwidth}
\setlength{\tabcolsep}{4pt}
  \centering
  \captionof{table}{\textbf{Ablation studies on decomposition network architecture.} We ablate the effectiveness of refinement work by studying 1-stage estimation and 2-stage estimation. The refinement network significantly improves the performance. 1-stage and 2-stage models are set to similar sizes by adjusting channel numbers, for fair comparison.
  }
  \label{tab:ablation}
  \resizebox{\linewidth}{!}{
  \begin{tabular}{lccc}
    \toprule
    Method & PSNR $\uparrow$ & SSIM $\uparrow$ & LPIPS $\downarrow$ \\
    \midrule
    1-stage & 24.47 & 0.912 & 0.074 \\
    2-stage & \bestscore{25.45} & \bestscore{0.933} & \bestscore{0.054} \\
    \bottomrule
  \end{tabular}
  }
\end{minipage}
\end{figure}

We present ablation studies on motion guidance and the multi-stage architecture. The experiments are conducted on B-Aist++ and the metrics are calculated using ground-truth motion guidance. Based on Fig.~\ref{fig:learning_curve}, introducing motion guidance in the training stage greatly improves convergence by eliminating directional ambiguity. Table~\ref{tab:ablation} shows that introducing a two-stage coarse-and-refinement pipeline while maintaining the model size may bring about \SI{1}{dB} gain.

\section{Conclusions}
\label{sec:conclusion}
In this work, we address the problem of recovering a sharp motion sequence from a motion-blurred image. We bring to light the issue of directional ambiguity and propose the first solution to this challenging problem by introducing motion guidance to train networks. The proposed method can adapt to blurry input of different modalities by using the corresponding interfaces including a multi-modal prediction network, motion from video, and user annotation. The motion sequences generated by our method are superior to existing methods in terms of quality and diversity.

\section*{Acknowledgement}
\label{sec:acknowledgement}
This work was supported by D-CORE Grant from Microsoft Research Asia, JSPS KAKENHI Grant Numbers 22H00529, and 20H05951, and JST, the establishment of university fellowships towards the creation of science technology innovation, Grant Number JPMJFS2108.

%
%
\bibliographystyle{splncs04}
\bibliography{egbib}

\appendix
\newpage
\section{Video Results}
We present the video results as \textcolor{red}{\texttt{results.mp4}}. We first show an overview of our multi-modal framework, which uses three different interfaces for the same sample to realize blur decomposition, including guidance from prediction network, motion of video and user annotation. Then, we show the video results including comparison samples on B-Aist++, GenBlur and real-world data.

\section{Quantization of Guidance}
\begin{figure}[t]
	\centering
	\includegraphics[width=.9\linewidth]{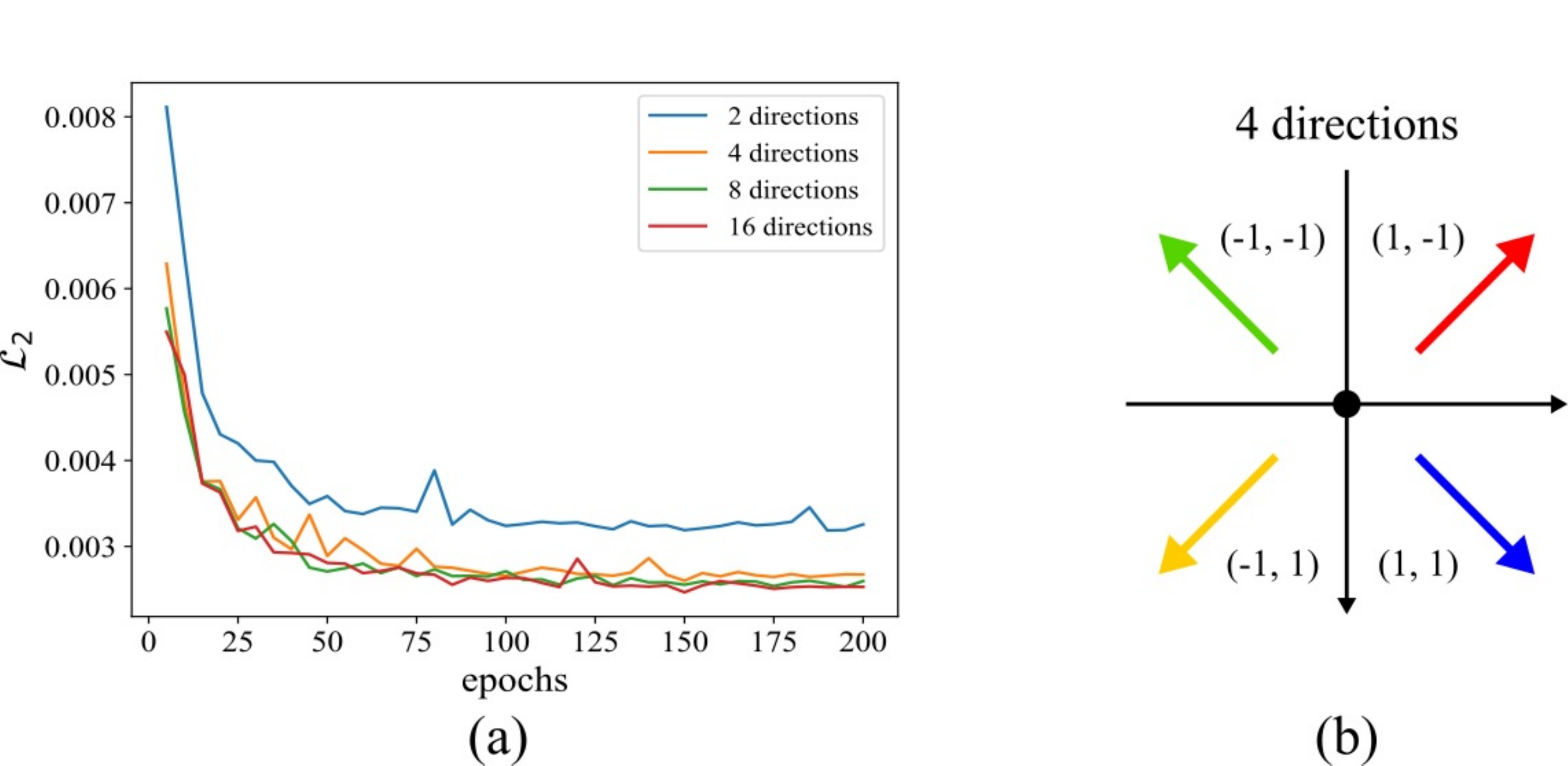}
	\caption{\textbf{Number of guidance directions.} (a) shows the validation loss curves of the model using guidance with 2, 4, 8, and 16 directions, respectively. (b) shows the guidance representation for the case of 4 directions that we used to present the visual results.}
	\label{fig:guidance_direction}
\end{figure}

As discussed in the manuscript, guidance with two directions is essentially sufficient to represent the forward and backward movement of each individual blurred region. We present the validation loss curves of the decomposition model using 2, 4, 8, 16 directions in Fig.~\ref{fig:guidance_direction} (a). This indicates that the use of 2 directions has been able to solve the problem of directional ambiguity to some extent. We find that 4 directions bring more gain than 2 directions, and then more directions such as 8 or 16 bring limited gain. Considering the ease of using guidance (\textit{e.g.}, user annotation) and the performance, we adopt 4 directions for other experiments.

We show the guidance representation for the case of 4 directions in Fig.~\ref{fig:guidance_direction} (b). Guidance is simplified as 5 classes for each pixel, including 4 quadrants and static, \textit{i.e.}, quadrant \Rmnum{1} $\left(+1, -1\right)$, quadrant \Rmnum{2} $\left(-1, -1\right)$, quadrant \Rmnum{3} $\left(-1, +1\right)$, quadrant \Rmnum{4} $\left(+1, +1\right)$, and origin $\left(0, 0\right)$, based on the motion direction. Regarding the case of 2, 8 and 16 directions, we use the similar 1-bit, 3-bit, 4-bit representation ($+1$ or $-1$ for each bit, all $0$ for static) to mark the directions.

\section{Network and Implementation Details}
\begin{figure}[!t]
	\centering
	\includegraphics[width=\linewidth]{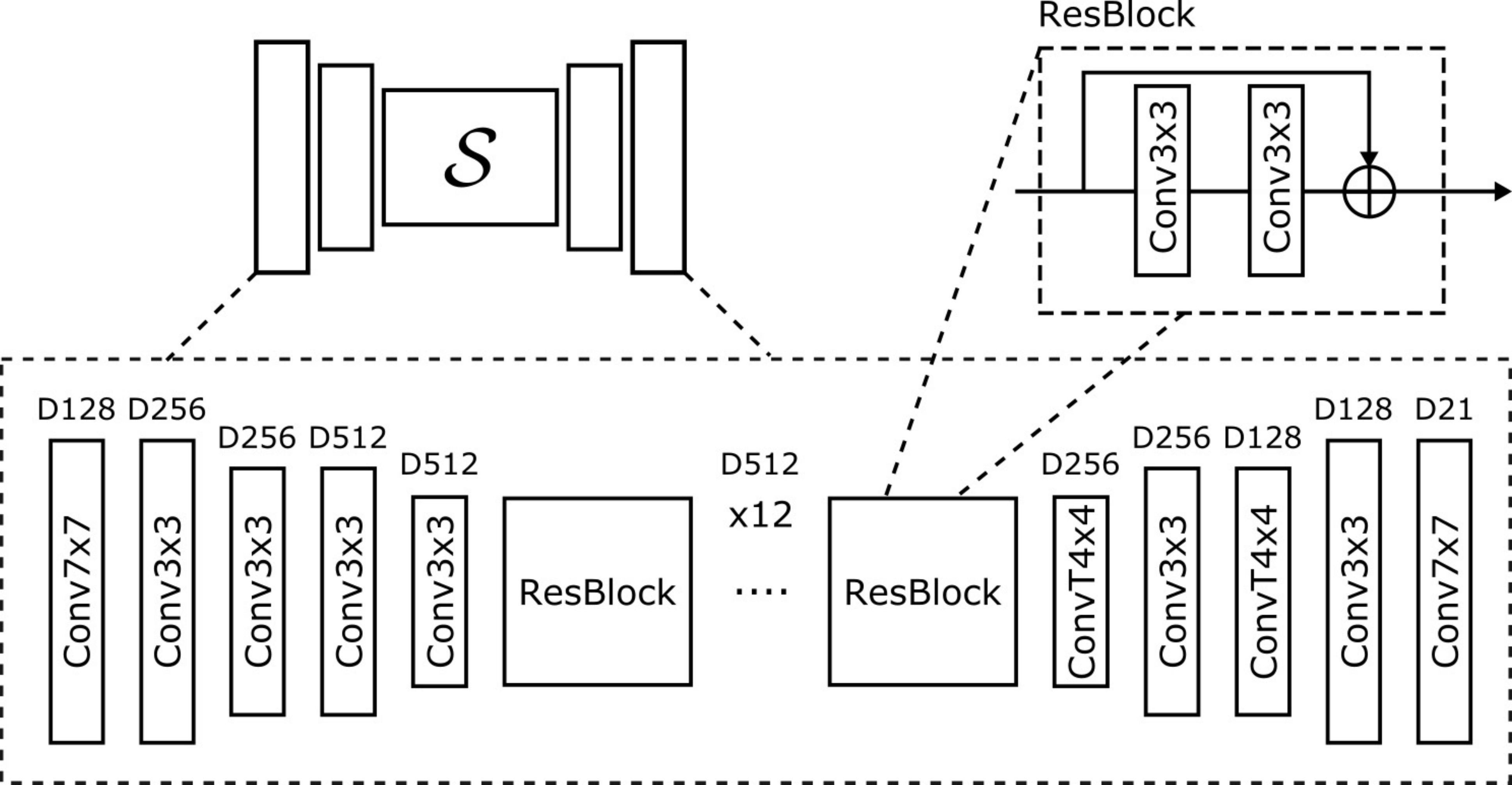}
	\caption{\textbf{The architecture details of a network stage in our 2-stage decomposition network.} The design of our stage network follows the commonly used encoder-decoder structure. It consists of two down-sampling 2d convolution steps, 12 ResBlocks~\cite{he2016deep} bottleneck, and two up-sampling 2d deconvolution steps. D\# denotes the number of channels after each convolution layer. The activation (ReLU~\cite{agarap2018deep}) and batch norm layers have been omitted in this figure for better clarity.}
	\label{fig:network_details}
\end{figure}

The architecture details of a stage network in our 2-stage decomposition network are illustrated in Fig.~\ref{fig:network_details}, which follows the design of~\cite{siarohin2019first}. The difference is that we replace the non-parametric up-sampling interpolation with a deconvolution layer. The architecture of our prediction network follows the cVAE-GAN of~\cite{zhu2017multimodal}. The difference is that we adopt a Gumble-Softmax~\cite{jang2016categorical} layer to the estimated guidance before calculating GAN loss. Please see the their \href{https://github.com/junyanz/BicycleGAN}{source code} for more details.

We use Pytorch~\cite{paszke2019pytorch} to implement our blur decomposition and guidance prediction networks. We trained the blur decomposer on GenBlur with a batch size of 16 for 500 epochs and on B-Aist++ with a batch size of 8 for 400 epochs. The learning rate is adjusted by a cosine scheduler with an initial learning rate of $2\times10^{-4}$ and $1\times10^{-4}$ for the two datasets, respectively. Flipping and random cropping ($256\times256$) are applied for data augmentation on GenBlur. For B-Aist++, cropping is applied to frame the person in the images, which are then resized to $192\times192$. To prevent overfitting, the dancers in the training set and the dancers in the test set of B-Aist++ are not intersected. The number of extracted frames is fixed to 7. Furthermore, motivated by the directional ambiguity, we augment each training sample by adding its inverse direction sample $(I_b, {G_{\text{inv}}}, \mathbf{I_{\text{inv}}})$ to the training data. Thanks to the ability of our network to handle directional ambiguity, including the inverse samples strengthens the dependency between $G$ and $\mathbf{I}$ and effectively increases the diversity of training samples. The setting of our guidance prediction network basically follows the cVAE-GAN of~\cite{zhu2017multimodal}. The reconstruction loss is replaced by a Cross-Entropy loss. Empirically, $\lambda_{1},\lambda_{2},\lambda_{3}$ are set as 0.1, 10, 0.1, respectively. For the comparisons, both the blurry image-based method~\cite{jin2018learning} and video-based method~\cite{shen2020blurry} are retrained on our datasets with the same number of epochs for fairness. We will release the datasets and source code to the community.

\section{Additional Qualitative Results}
\begin{figure}[!t]
	\centering
	\includegraphics[width=\linewidth]{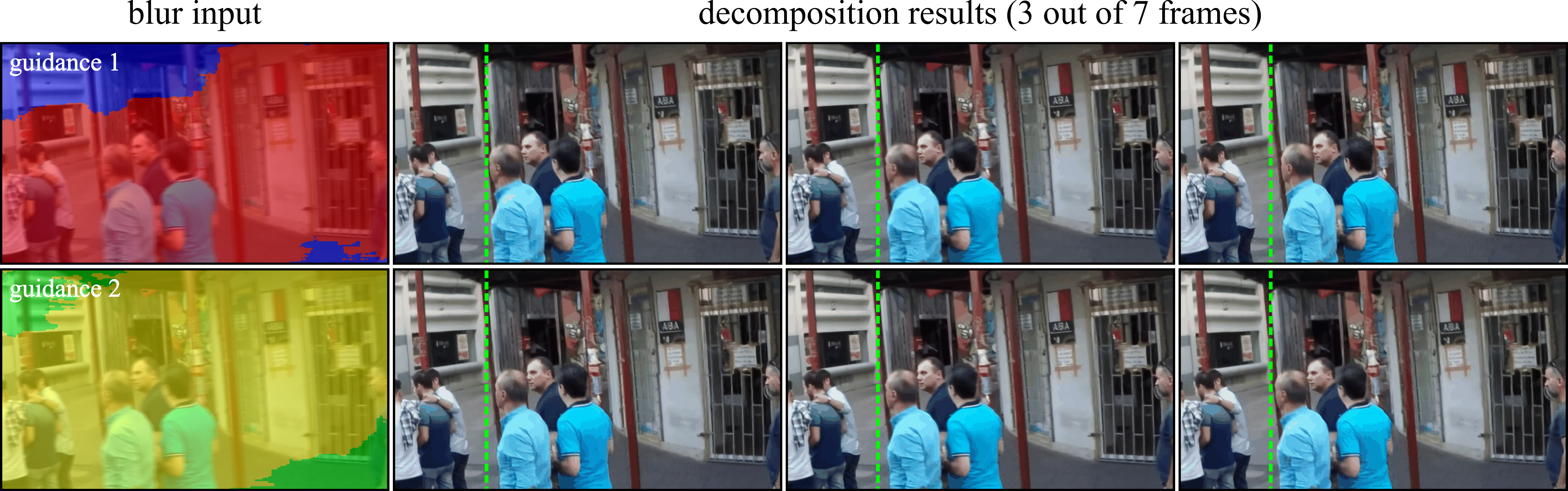}
	\caption{\textbf{Multi-modal predictions on a sample of GenBlur dataset.}}
	\label{fig:genblur}
\end{figure}

We additionally provide an example of multi-modal predictions on the GenBlur dataset, as illustrated in Fig.~\ref{fig:genblur}. Besides, we provide our guidance predictions and blur decomposition results for real-world data, along with the results of~\cite{jin2018learning} in Fig.~\ref{fig:real}.

\begin{figure}[!t]
	\centering
	\includegraphics[width=\linewidth]{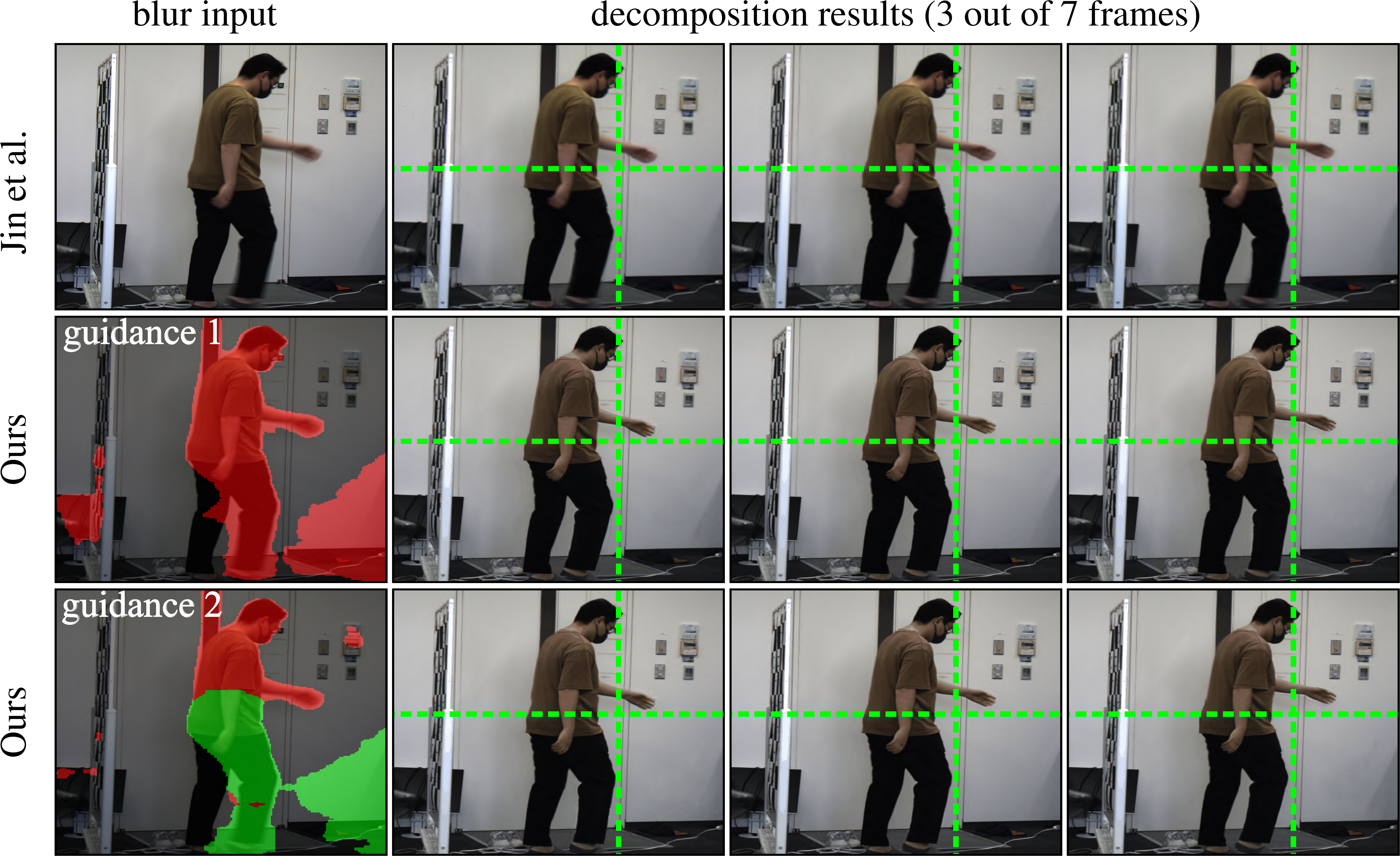}
	\caption{\textbf{Real-world evaluations with the guidance prediction network.}}
	\label{fig:real}
\end{figure}

\begin{figure}[!t]
	\centering
	\includegraphics[width=\linewidth]{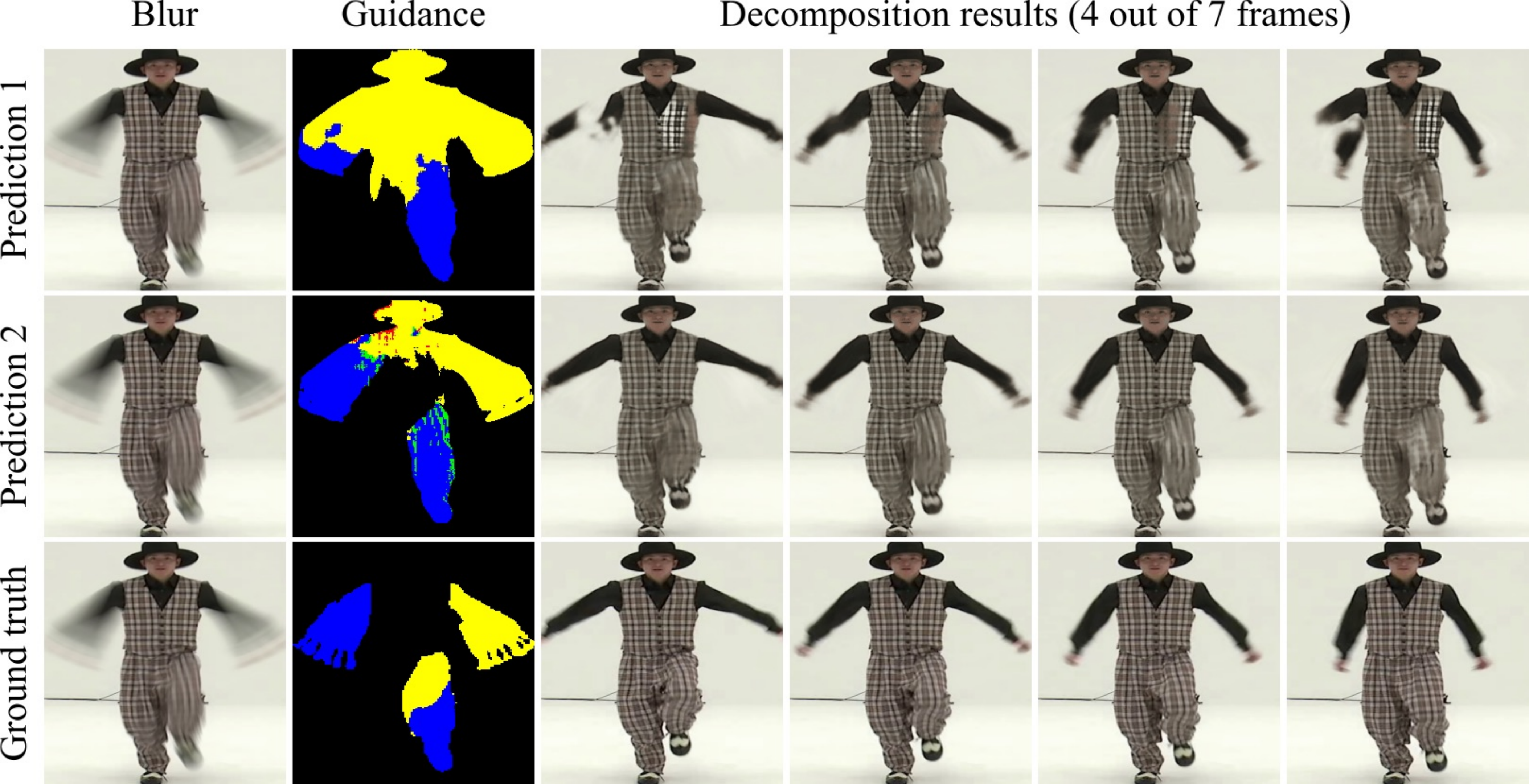}
	\caption{\textbf{Failure case of our method.} 
		The guidance prediction network may produce wrong guidance in the face of extremely severe blur. In addition, it is very challenging for the decomposition network to restore the details of complex texture in the case of severe blur.}
	\label{fig:failure}
\end{figure}

\section{Failure Case}
We show the failure cases of our approach in the face of severe blur, as illustrated in Fig.~\ref{fig:failure}. Severe blur can cause difficulties for the guidance prediction network and thus may produce incorrect guidance like the first row (Note the left arm) of Fig.~\ref{fig:failure}. In addition, it is very challenging for the decomposition network to restore the details of complex texture (Note the stripes on the right pants) in the case of severe blur. However, this does not affect the fact that the proposed guidance can help resolve the problem of directional ambiguity.

\section{Limitations}
\label{sec:limitations}
The current simple form of motion guidance cannot handle the case with large and complex motion blur. Using a more elaborate guidance that takes into account different motion intensities may improve this problem. Regarding the training data, we did not consider the camera noise when synthesizing the data. This may impair the performance of the model on real-world data to some extent. Using a beam-splitter acquisition system~\cite{rim2020real,zhong2020efficient} to collect a real-world blur decomposition dataset will be a promising future work.
\end{document}